\documentclass[10pt,twocolumn,letterpaper]{article}

\usepackage{cvpr}              

\usepackage{graphicx}
\usepackage{amsmath}
\usepackage{amssymb}
\usepackage{booktabs}

\usepackage{comment}
\usepackage{tabu}
\usepackage{rotating}
\usepackage{epstopdf}
\usepackage{balance}
\usepackage{multirow}
\usepackage{textcomp}
\usepackage{bm}

\usepackage{fancyhdr}
\usepackage{color}
\usepackage{amsfonts,amscd}
\usepackage{array}
\usepackage{booktabs}
\usepackage{graphics}
\usepackage{url}
\usepackage{psfrag,paralist}
\usepackage{multirow}
\usepackage{tabularx}
\usepackage{enumitem}
\usepackage[table]{xcolor}
\usepackage{bbm}
\usepackage{cancel}
\definecolor{lightgray}{gray}{.9}

\usepackage[pagebackref,breaklinks,colorlinks,bookmarks=false,citecolor=blue]{hyperref}

\usepackage[capitalize]{cleveref}
\crefname{section}{Sec.}{Secs.}
\Crefname{section}{Section}{Sections}
\Crefname{table}{Table}{Tables}
\crefname{table}{Tab.}{Tabs.}

\def\cvprPaperID{9787} 
\def\confName{CVPR}
\def\confYear{2022}

\graphicspath{{figure/}}

\makeatletter
\DeclareRobustCommand\onedot{\futurelet\@let@token\@onedot}

\newcommand{\red}{\color{red}}
\newcommand{\blue}{\color{blue}}
\newlength\figsep
\newlength\eqsep
\newcommand{\rb}[2]{\raisebox{#1 pt}{#2}}

\begin{document}

\title{Alignment-Uniformity aware Representation Learning \\
for Zero-shot Video Classification}

\author{Shi Pu$^1$\thanks{These authors contributed equally.}
\quad Kaili Zhao$^{\text{2}*}$
\quad Mao Zheng$^1$\\
\hspace{1.2cm}$^1$Tencent AI Platform Department \hspace{1.3cm}
$^2$Beijing University of Posts and Telecom.
\\
\hspace{-0.5cm}\texttt{\{shipu, moonzheng\}@tencent.com \hspace{0.6cm} kailizhao@bupt.edu.cn}\\
\vspace{-2.5ex}
}

\maketitle

\begin{abstract}
Most methods tackle zero-shot video classification by aligning visual-semantic representations within seen classes, which limits generalization to unseen classes.
To enhance model generalizability, this paper presents an end-to-end framework that preserves alignment and uniformity properties for representations on both seen and unseen classes.
Specifically, we formulate a supervised contrastive loss to simultaneously align visual-semantic features (\ie, alignment) and encourage the learned features to distribute uniformly (\ie, uniformity).
Unlike existing methods that only consider the alignment, we propose uniformity to preserve maximal-info of existing features, which improves the probability that unobserved features fall around observed data. 
Further, we synthesize features of unseen classes by proposing a class generator that interpolates and extrapolates the features of seen classes. 
Besides, we introduce two metrics, closeness and dispersion, to quantify the two properties and serve as new measurements of model generalizability.
Experiments show that our method significantly outperforms SoTA by relative improvements of 28.1\% on UCF101 and 27.0\% on HMDB51.
Code is available\footnote{\url{https://github.com/ShipuLoveMili/CVPR2022-AURL}}.

\end{abstract}

\section{Introduction}
Mimicking human capability to recognize things never seen before, zero-shot video classification (ZSVC) only trains models on videos of seen classes and makes predictions on unobserved ones \cite{liu2011recognizing,vgan,multi,mettes2017spatial,mettes2021object,gao2020learning,kim2021daszl,zhu2018towards}.
Correspondingly, existing ZSVC models map visual and semantic features into a unified representation, and hope the association can be generalized to unseen classes \cite{zhu2018towards,hahn2019action2vec,bishay2019tarn,brattoli2020rethinking,qiu2021boosting,chen2021elaborative}.
However, these methods learn associated representations within limited classes, thus facing the following two critical problems \cite{fu2015transductive,gao2020learning}:
(1) semantic-gap: manifolds inconsistency between visual and semantics features, and 
(2) domain-shift: the representations learned from training sets are biased when applied to the target sets due to disjoint classes between two groups.
In ZSVC, these two problems cause side effects on model generalizability.

\begin{figure}
    \centering
    \includegraphics[width=0.45\textwidth]{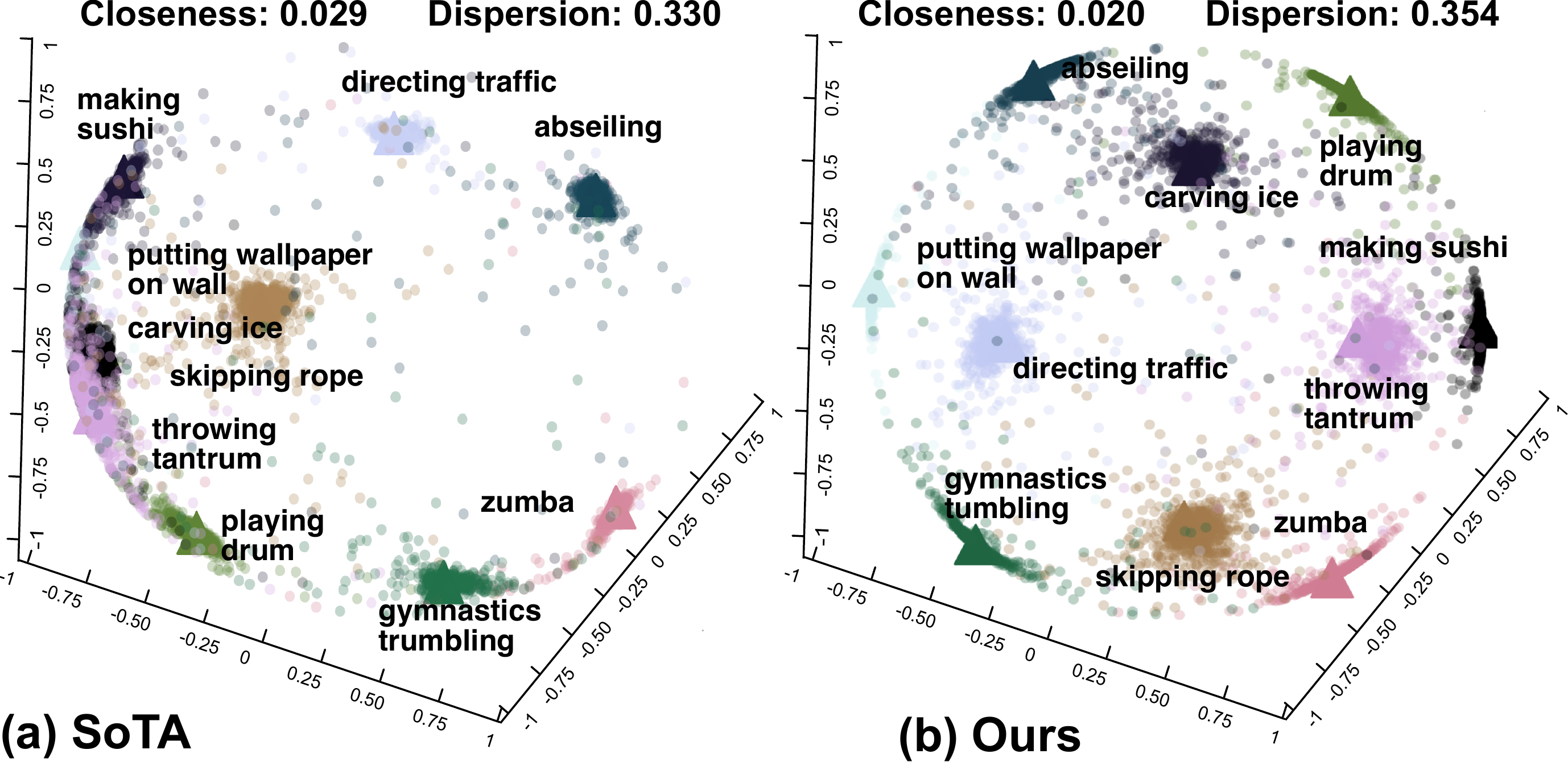}
    \caption{
             {\bf Visual-semantic representations:} Comparisons of the learned representations between the SoTA \cite{brattoli2020rethinking} and our method.
             \protect\scalebox{1}{\textbullet} and \protect$\bigtriangleup$ represent visual and semantic features separately; colors are for different classes.
             Besides, we use two metrics to quantify feature qualities on alignment (closeness$\downarrow$ better) and uniformity (dispersion$\uparrow$ better).
             We observe that ours show better closeness within classes and more separations among semantic clusters.}
    \label{fig:fig1}
    \vspace{\figsep}
\end{figure}

Reviewing the literature, we observe that most methods focus on tackling the semantic-gap by learning alignment-aware representations, which ensure visual and semantic features of the same class close.
To improve the alignment, MSE loss \cite{brattoli2020rethinking}, ranking loss \cite{hahn2019action2vec}, and center loss \cite{gao2020learning} are commonly used to optimize the similarity between visual and semantic features. 
Apart from the loss, improvements for alignment are attributed mainly to the designs of architectures.
For instance, \cite{jain2015objects2action,gao2020learning,mettes2021object} first project global visual features to local object attributes, then optimize similarity between the attributes and final semantics.
In contrast, URL \cite{zhu2018towards}, Action2Vec \cite{hahn2019action2vec}, and TARN \cite{bishay2019tarn} directly align visual and final semantic features, which are improved via attention modules.
Since video features are hard to learn, the above methods utilize pre-trained models to extract visual features.
The recent model \cite{brattoli2020rethinking} benefits from the efficient R(2+1)D module \cite{tran2018closer} in video classification and achieves the state-of-the-art (SoTA) results in ZSVC.
However, the SoTA \cite{brattoli2020rethinking} neglects to learn semantic features; thus, it is still not a true end-to-end (e2e) framework for visual-semantic feature learning.
We claim that e2e is critical for alignment since fixed visual/semantic features will bring obstacles to adjusting one to approach another.

Noteworthily, the latest MUFI \cite{qiu2021boosting} and ER \cite{chen2021elaborative} get down to addressing the domain-shift problem by involving more semantic information, thus consuming extra resources.
In particular, MUFI \cite{qiu2021boosting} augments semantics by training multi-stream models on multiple datasets.
ER \cite{chen2021elaborative} expands class names by annotating amount of augmented words crawled from the website.
Freeing complex models or additional annotations, we will design a compact model that preserves maximal semantic info of existing classes while synthesizing features of unseen classes.

To tackle the two problems with one stone, we present an end-to-end framework that jointly preserves alignment and uniformity properties for representations on both seen and unseen classes.
Here, alignment ensures closeness of visual-semantic features; uniformity encourages the features to distribute uniformly (maximal-info preserving), which improves the possibility that unseen features stand around seen features, mitigating the domain-shift implicitly.
Specifically, we formulate a supervised contrastive loss as a combination of two separate terms: one regularizes alignment of features within classes, and the other guides uniformity between semantic clusters.
To alleviate the domain-shift explicitly, we generate new features of unseen synthetic classes by our class generator that interpolates and extrapolates features of seen classes. 
In addition, we introduce closeness and dispersion scores to quantify the two properties and provide new measurements of model generalizability.
Fig.~\ref{fig:fig1} illustrates the representations of our method and the SoTA alternative \cite{brattoli2020rethinking}.
We train the two models on ten classes sampled from Kinetics-700 \cite{carreira2017quo} and map features on 3D hyperspheres.
We observe that our representation shows better closeness within classes and preserves more dispersion between semantic clusters.
Experiments validate that our method significantly outperforms SoTA by relative improvements of 28.1\% on UCF101 and 27.0\% on HMDB51.


\section{Related Work}
\label{sec:relatedwork}

{\bf Supervised video classification (SVC):}
SVC tackles general classes initially (\eg, YouTube-8M dataset \cite{abu2016youtube}), then specific to action recognition recently (\eg, large-scale Kinetics-700 dataset\cite{carreira2017quo}).
Learning temporal features is the main task of SVC.
In the beginning, video features are generated via NetVLAD \cite{miech2017learnable,lin2018nextvlad} that fuses static features of multiple frames.
Then, temporal/motion features of videos are optimized directly. 
We categorize the methods into two-stream 2D-CNN and 3D-CNN based.
Two-stream models \cite{simonyan2014two,wang2016temporal,lin2019tsm} extract spatial and temporal features by performing separate 2D-CNN modules.
\cite{jiang2019stm,liu2020teinet} extracts motion features by computing 2D-CNN features' difference between neighboring frames.
Furthermore, \cite{tran2015learning} proposes C3D to fuse spatial and temporal features via an independent 3D-CNN module.
Even C3D helps achieve promising results \cite{feichtenhofer2019slowfast,sudhakaran2020gate}, its large parameters bring burdens to model optimization.
Instead, I3D \cite{carreira2017quo} and P3D \cite{qiu2017learning} design 3D-CNN-like modules by combining 1D temporal and 2D spatial filters. 
Recently, a more efficient R(2+1)D module \cite{tran2018closer} has been widely used, which includes a pseudo-3D kernel (2D spatial + 1D temporal) in residual networks.
In this paper, we apply our model in action recognition and perform R(2+1)D for better spatial-temporal feature extraction.
\begin{figure*}
    \centering
    \includegraphics[width=0.9\textwidth]{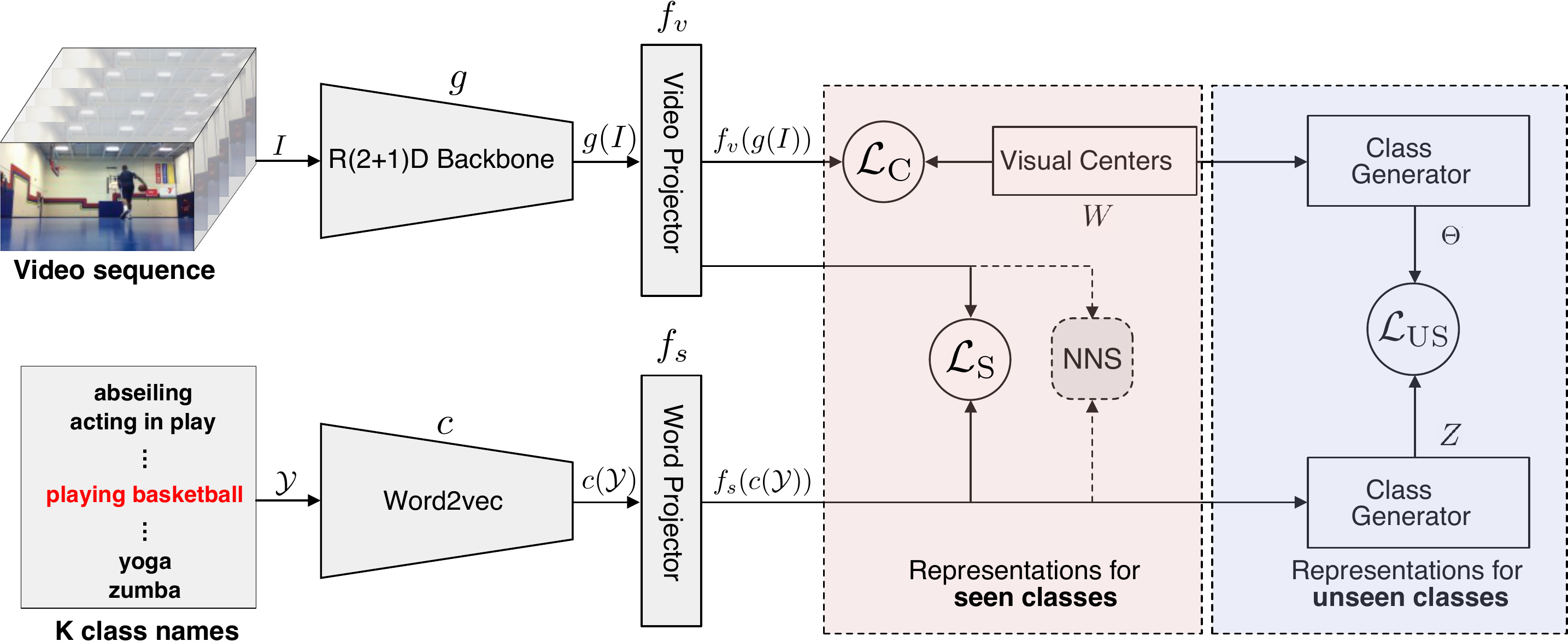}
    \caption{
            {\bf Architecture of AURL:} 
            From left to right, we map a video sequence $I$ and the class name set $\mathcal{Y}$ to a unified representation ($f_v(g(I))$, $f_s(c(\mathcal{Y}))$).
            During training, to learn representations of seen classes, we introduce $\mathcal{L}_\text{S}$ to preserve {\em alignment} and {\em uniformity} properties.
            For synthetic unseen classes, we introduce $\mathcal{L}_\text{US}$ to learn the two properties on synthetic visual-semantic features ($\Theta, Z$).
            To synthesize features of unseen classes, we first utilize $\mathcal{L}_\text{C}$ to learn {\em visual centers} $W$, then propose {\em Class Generator} to transform $W$ and existing semantics $f_s(c(\mathcal{Y}))$ into the representation ($\Theta, Z$).
            During inference, we perform an NNS strategy to obtain the final class.
            }
    \label{fig:fig2}
    \vspace{\figsep}
\end{figure*}

{\bf Zero-shot video classification (ZSVC):}
Existing ZSVC methods align visual and semantic features on a unified representation and hope the alignment can be generalized to unseen classes.
Most methods design various frameworks to optimize the alignment.
Similar to zero-shot image classification \cite{xian2018zero}, some methods \cite{liu2011recognizing,gan2016learning,jain2015objects2action,mettes2021object,kim2021daszl} learn video attributes first, then design stage-wise framework.
Given input videos, \cite{liu2011recognizing,gan2016learning} learn classifiers for video attributes, then compare the predicted attributes and final class names.
However, they cost intensive annotations of video attributes.
Instead, \cite{jain2015objects2action,mettes2021object,kim2021daszl} utilize pre-trained object detectors to determine object-level class names, then compute similarities between object-level and final class names. 
Recent work directly computes the similarity between visual and semantic features, and their contributions focus on enhancing visual features.
URL \cite{zhu2018towards}, TARN \cite{bishay2019tarn} and Action2Vec \cite{hahn2019action2vec} extract spatial-temporal features using a pre-trained C3D and then improve the features via attention modules.
The latest model \cite{brattoli2020rethinking} learns visual embeddings by an efficient R(2+1)D module and achieves SoTA results.
However, the above methods are not true end-to-end (e2e) models because those utilize Word2vec \cite{mikolov2013efficient} to extract semantic features.
We will justify that lacking e2e learning weakens the alignment since fixed visual/semantic features will bring obstacles to adjusting one to approach another.
Except for the above designs, MSE loss \cite{brattoli2020rethinking}, ranking loss \cite{hahn2019action2vec}, and center loss \cite{gao2020learning} are commonly used to regularize the alignment of features.
In this paper, we propose a true e2e framework and formulate a supervised contrastive loss, which first considers both alignment and uniformity properties in ZSVC.

{\bf Representation learning:}
In self-supervised and zero-shot learning, representation learning learns features of observed data, which can extract helpful info when applied to downstream tasks.
In self-supervised learning, given pairs of positive and negative images/videos, contrastive learning regularizes representations where positives stand close while negatives keep apart.
The pioneering work, SimCLR \cite{chen2020simple} utilizes data augmentation to generate positive instances and maintains a large batch for choosing relatively enough negatives.
\cite{qian2021spatiotemporal} applies SimCLR to the video domain.
To save memory, MoCo \cite{he2020momentum} presents momentum update to cache a large number of negative instances, then \cite{kuang2021video, pan2021videomoco} extends MoCo to video understanding.
\cite{zhu2021improving} introduces feature transformation on existing samples to obtain broader and diversified positives and negatives, thus enhancing discrimination.
However, the above models learn instance pairs from the same domain, \eg, images or videos.
Instead, CLIP \cite{radford2021learning} exploits features of two domains (\ie, images and texts) guided by a contrastive loss.
Motivated by CLIP, the latest models in ZSVC, MUFI \cite{qiu2021boosting} and ER \cite{chen2021elaborative}, extend the self-supervised contrastive loss to a fully-supervised loss that contrasts visual and semantic features.
However, MUFI and ER neglect the difference and similarities of the self-supervised and supervised contrastive losses.
Considering alignment and uniformity properties, we build connections between the two losses and analyze the advantages of the supervised loss.
Besides, MUFI and ER both require extra resources to improve model generalizability.
We propose a compact model using a class generator to explicitly synthesize new features of synthetic unseen classes.


\section{Alignment-Uniformity aware Representation Learning (AURL)}
This section describes Alignment-Uniformity aware Representation Learning (AURL) involving a unified architecture, loss functions, class generator, and two novel metrics, followed by its training and inference strategy, and then discusses similarities and differences against alternatives.

\subsection{Architecture}
Fig. \ref{fig:fig2} shows the AURL architecture.
Given complete $K$ class names $\mathcal{Y}=\{y_1, \dots, y_K\}$, and an input video $I$ of class $y_i\in \mathcal{Y}$ (\eg, playing basketball), we by end-to-end learn visual and semantic embeddings.
We introduce R(2+1)D \cite{tran2018closer} as the backbone to generate visual features $g(I)$, and utilize a video projector $f_v$ to implement 3-layer MLP projection (2 fc+bn+ReLU and 1 fc+bn), thus obtain visual embeddings $f_v(g(I))\in\mathbb{R}^{d}$.
Parallelly, we perform Word2vec $c$ \cite{mikolov2013efficient} to extract the initial word embeddings $c(\mathcal{Y})$, then learn semantic embeddings $f_s(c(\mathcal{Y}))\in\mathbb{R}^{K\times d}$ by a word projector $f_s$ that has one fc (\#node=512) and the 3-layer MLP projection. 
For convenience, we note $f_v(g(I))$ and $f_s(c(y_i))$ of $i$-th class as $v_{y_i}$ and $s_{y_i}$ for the below discussions.
Compared with SoTA methods \cite{brattoli2020rethinking,qiu2021boosting} that only learn video parts, our AURL end-to-end trains the backbone, video and word projectors, providing more feature flexibility under the regularization of loss functions.  

\subsection{Alignment-uniformity aware Loss}
{\em Can alignment and uniformity properties be preserved in supervised contrastive loss?}
For self-supervised learning, \cite{chen2020intriguing} claims that contrastive loss \cite{chen2020simple,qian2021spatiotemporal} (see Eq.~\ref{eq:self} and {\em supplementary}) preserves alignment and uniformity properties.
Alignment indicates that positive samples should be mapped to nearby features and thus be invariant to unneeded noises.
Uniformity \cite{wang2020understanding} means feature vectors should be roughly uniformly distributed on the unit hypersphere, thus bringing better generalization to downstream tasks.

\noindent
\begin{align}
\mathcal{L}^{self} \!=& \!-\! \log[\frac{\exp{[\lambda \text{sim}(f, f_+)]}}{\sum_{f_- \in \mathcal{N} }\exp{[\lambda \text{sim}(f, f_-)]}}]. 
\label{eq:self}
\vspace{\eqsep}
\end{align}
Here, $(f, f_+), (f, f_-)$ are positive and negative pairs of images/videos, $\mathcal{N}$ is negative set; sim means a similarity function (thus in [-1, +1]); and $\lambda$ is a temperature parameter.

Closeness in alignment and maximal-info preserving in uniformity are also essential properties of the unified representation learning in ZSVC.
However, existing work mainly focuses on the alignment of visual-semantic features \cite{zhu2018towards,bishay2019tarn,hahn2019action2vec,brattoli2020rethinking}, the uniformity that improves generalization has not been discussed yet.
Here, by leveraging video labels, we formulate a supervised contrastive loss as the combination of alignment and uniformity terms:
\begin{align}
\label{eq:sup}
\mathcal{L}^{sup}\! =& \!-\! \log[\frac{\exp{[\lambda \text{sim}(v_{y_i}, s_{y_i})]}}{\sum_{y_j \in \mathcal{Y} }\exp{[\lambda \text{sim}(v_{y_i}, s_{y_j})]}}],\\ \nonumber
                            \!=&\lambda \text{SP}_\lambda [
                            \underbrace{-\text{sim}(v_{y_i}, s_{y_i})}_\text{alignment} \!+ 
                            \frac{1}{\lambda} \underbrace{\text{LSE}(\lambda\text{sim}(v_{y_i}, s_{y_j})_{y_j \in \mathcal{Y}\setminus y_i})}_\text{uniformity} ],\\ \nonumber 
&\text{where,}\quad \;\text{SP}_\lambda(x)\! =\! \frac{1}{\lambda}\log(1\!+\!\exp(\lambda x)),\\ \nonumber
& \quad\quad\quad\quad \text{LSE}(x) \!=\! \log(\sum_{x\in\mathcal{X}}\exp(x)).
\vspace{\eqsep}
\end{align}
$v_{y_i}$ and $s_{y_i}$ are visual and semantic features of class $y_i$, and the complete class set is $\mathcal{Y}$. $\text{SP}_\lambda$ means the SoftPlus function and LSE is LogSumExp. 
Since Eq.~\ref{eq:sup} favors $\text{sim}(v_{y_i}, s_{y_i})$ larger, visual and semantic features of the same class will be aligned.
The uniformity term tends to maximize the distances between features of different classes using a LogSumExp function, thus spreading features as much as possible.
To sum up, our $\mathcal{L}^{sup}$ preserves the alignment and uniformity properties simultaneously.

{\bf $\mathcal{L}^{sup}$ performs better:} Comparing $\mathcal{L}^{sup}$ with $\mathcal{L}^{self}$, we observe that $\mathcal{L}^{sup}$ includes positive pair $\text{sim}(v_{y_i}, s_{y_i})$ in the denominator.
Even recent work MUFI \cite{qiu2021boosting} and ER \cite{chen2021elaborative} also utilize supervised contrastive loss, not only do they neglect the alignment and uniformity properties, but also miss the similarity and difference between the two losses.
Here, we show that $\mathcal{L}^{sup}$ maintains advantages of both $\mathcal{L}^{self}$ and triplet loss \cite{schroff2015facenet}.
We derive upper bounds of $\mathcal{L}^{sup}$ and $\mathcal{L}^{self}$ as follows (the full derivation in {\em supplementary}):
\begin{align}
\label{eq:bound}
    \mathcal{L}^{self} \!&\leq\!\lambda(\text{sim}_\text{max}\!-\!\text{sim}(v_{y_i}, s_{y_i})+\frac{\log(K\!-\!1)}{\lambda}),\\\nonumber
    \mathcal{L}^{sup} \!&\leq\! \lambda\max[\text{sim}_\text{max}\!-\!\text{sim}(v_{y_i}, s_{y_i})\!+\!\frac{\log(K\!-\!1)}{\lambda}, 0] \!+\!\log(2), 
    \vspace{\eqsep}
\end{align}
where $K$ is the number of classes, $\text{sim}_\text{max}$ is the maximal similarity among all negative pairs ($\text{sim}_\text{max} = \max_{y_j \in \mathcal{Y}\setminus{y}}\text{sim}(v_{y_i},s_{y_j})$).
For a fair comparison, we also reformulate $\mathcal{L}^{self}$ with class labels and obtain its upper bound in Eq.~\ref{eq:bound}.
With the upper bounds, we summarize the advantages of $\mathcal{L}^{sup}$ as follows:
\begin{compactenum}
\item When $\frac{\log (K-1)}{\lambda}\geq 2$, the two upper bounds will be similar. At this time, $\mathcal{L}^{sup}$ performs as well as $\mathcal{L}^{self}$ in a representation learning task.
\item When $0\leq\frac{\log(K-1)}{\lambda}\textless 2$, the upper bound of $\mathcal{L}^{sup}$ has a similar form as triplet loss \cite{schroff2015facenet} that facilitates intrinsic ability to perform hard positive/negative mining.
\item $\mathcal{L}^{sup}$ preserves the summation over all negatives in the denominator, thus improving discrimination among classes \cite{sohn2016improved}, which has the same motivation with contrastive learning that makes the embedding distribution uniform by increasing the number of negatives \cite{chen2020simple}.
\end{compactenum}
In this paper, we take advantage of $\mathcal{L}^{sup}$ to regularize the representations of both seen and synthetic unseen classes. 

\begin{figure*}
    \centering
    \includegraphics[width=0.95\textwidth]{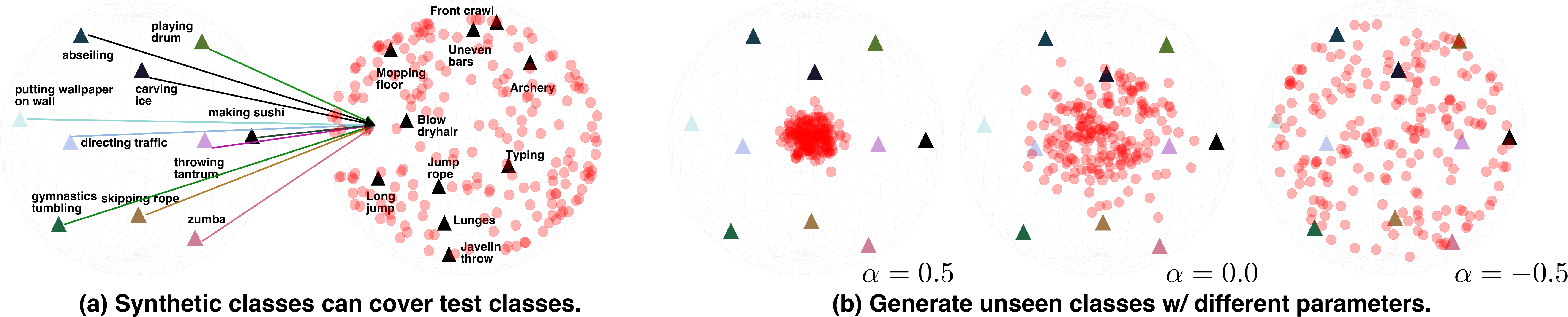}
    \caption{
    {\bf Illustration of Class Generator:}
    (a) Synthetic classes \protect\scalebox{1}{\color{red}{\textbullet}} are generated by linearly combining features of seen classes (\protect\scalebox{0.75}{$\triangle$} with colors); $\blacktriangle$ means test classes.
    (b) Feature transformation w/ different $\alpha$ synthesizes semantics covering various size of regions.}
    \label{fig:generator}
    \vspace{\figsep}
\end{figure*}

{\bf $\mathcal{L}^{sup}$ for seen and unseen classes:}
We learn $\mathcal{L}^{sup}$ for both seen and unseen classes (see $\mathcal{L}_\text{contrast}$ in Eq.~\ref{eq:contrast} where we utilize $\cos$ as a cosine function and map features on the hypersphere.).
Specifically, we learn $\mathcal{L}_\text{S}$ on visual-semantic features (\ie, $v_{y_i}$ and $s_{y_i}\in\mathbb{R}^{1\times d}, y_i\in\mathcal{Y}$) for seen classes set $\mathcal{Y}$.
From the formulation of $\mathcal{L}_\text{S}$, we jointly align features of the same class and introduce uniformity that encourages semantic clusters to spread as much as possible, improving the possibility that features of unseen classes fall around existing ones.
To offer effective positive/negative visual and semantic pairs that enhance the feature embedding \cite{zhu2021improving}, we propose a {\em class generator} to generate visual and semantic features of synthetic classes $\mathcal{U}$, which are considered as ``unseen classes" in comparison with seen classes $\mathcal{Y}$.
To retain the {\em alignment-uniformity} properties, we utilize $\mathcal{L}_\text{US}$ to regularize visual and semantic features of $K_u$ unseen classes (\ie, the synthetic features $\Theta $ and $Z \in \mathbb{R}^{K_u\times d}$).
\begin{align}
\label{eq:contrast}
    \mathcal{L}_\text{contrast} \!= &\mathcal{L}_\text{S} + \mathcal{L}_\text{US}\\\nonumber
    =&\!-\! \log[\frac{\exp{[\lambda \text{cos}(v_{y_i}, s_{y_i})]}}{\sum_{y_j \in \mathcal{Y} }\exp{[\lambda \text{cos}(v_{y_i}, s_{y_j})]}}] + \\ \nonumber 
    & \frac{1}{K_u}\sum_{u_i\in\mathcal{U}}\!-\! \log[\frac{\exp{[\lambda \text{cos}(\Theta_{u_i}, Z_{u_i})]}}{\sum_{u_j \in \mathcal{U} }\exp{[\lambda \text{cos}(\Theta_{u_i}, Z_{u_j})]}}].
    \vspace{\eqsep}
\end{align}

\subsection{Class Generator}
To synthesize visual and semantic pairs ($\Theta, Z$), we propose a class generator that applies a uniformly sampled linear transformation to all pairs of visual/semantic features of seen classes.
Especially, instead of using single visual feature as the transformed features, we select representative ``visual centers" learned from a supervised video classifier that interprets the parameter matrix of fc layer as the centers, as commonly used in \cite{wang2017normface,wang2018additive,du2019explicit,press2016using}.
We propose to exploit $\mathcal{L}_\text{C}$ as the classification loss, which is an angular softmax loss, helping push all visual features towards their visual centers on the unit hyperspher \cite{Deepcosine}. 
\begin{align}
    \label{eq:visual}
    \mathcal{L}_\text{C} = -\log \frac{\exp[\lambda\cos (v_{y_i}, w_{y_i})]}{\sum_{y_j \in \mathcal{Y}}\exp[\lambda\cos(v_{y_i}, w_{y_j})]}.
    \vspace{\eqsep}
\end{align}
Here, $v_{y_i}$ and $w_{y_i}$ indicate a single visual feature and the learned visual center, respectively.  
With the visual centers $w_{y_i}$ and the corresponding semantic features $s_{y_i}$, we inter-and extra-polates (\ie, linearly combine) these features to fill in incomplete class points on the hypersphere:
\begin{align}
\vspace{-10cm}
\label{eq:generator}
\Theta               &= M \times \text{norm}(W),\\\nonumber
Z                    &= M \times \text{norm}(s_{\mathcal{Y}}),\\\nonumber
\text{norm}(W)      &= [\frac{w_{y_1}}{\left\|w_{y_1}\right\|},\dots,\frac{w_{y_D}}{\left\|w_{y_D}\right\|}],\\\nonumber
\text{norm}(s_{\mathcal{Y}})  &=[\frac{s_{y_1}}{\left\|s_{y_1}\right\|},\dots,\frac{s_{y_{D}}}{\left\|s_{y_{D}}\right\|}].
\vspace{\eqsep}
\end{align}
Here, $\Theta$ and $Z$ $\in\mathbb{R}^{K_u\times d}$ separately represent the synthetic visual centers and semantic features of unseen classes;
$K_u$ represents the number of unseen classes and $d$ is the feature dimension.
Besides, we apply normalization on $W$ and $s_\mathcal{Y}$ (both are $D\times d$ matrix) because learning on a unit hypersphere helps model optimization \cite{wang2020understanding};
$D$ is a hyper-parameter that means how many classes are sampled for unseen-class generator.
The matrix $M\in\mathbb{R}^{K_u\times D}$ is used for inter- and extra-polations, whose elements are randomly sampled from a uniform distribution $U(\alpha, 1)$, and $-1\leq\alpha\textless 1$.
$\alpha$ is another hyper-parameter that controls the distributed range of the synthetic points.

It is worth noting that the settings of hyper-parameters $D$ and $\alpha$ are non-trivial.
For $D$, we prefer $D \geq d$, \ie, the number of seen classes should be larger than the dimension of a hypersphere.
Because for a full rank matrix $W$, a linear combination of the column vector of $W$ can express any vector on the transformed space.
We aim to generate as diverse unseen classes as possible to improve the possibility that the synthesized points can cover the classes in the test set. 
Thus, in experiments, we will select features of {\em all} seen classes for feature transformation. 
For $\alpha$, we choose positive values for interpolation where the synthetic clusters locate inside of seen points (see Fig.~\ref{fig:generator}(b)), and gradually enlarge the cluster regions by decreasing $\alpha$ where the negative value is for extrapolation.
Fig.~\ref{fig:generator} (a) illustrates our {\em Class Generator} with $D=10, d=3, \alpha=-1$ on the Kinetics-700 dataset \cite{kay2017kinetics}.
We can see our transformation not only provides unseen classes but also approaches test classes (\eg, UCF101 dataset \cite{soomro2012dataset}).

\subsection{Closeness and Dispersion}
To quantify the alignment and uniformity, we introduce two metrics: closeness and dispersion. 
Closeness measures the mean distance of features within the same class, reflecting the alignment of visual and semantic features.
\begin{align}
    \label{eq:closeness}
    \text{Closeness} = \frac{1}{K}\sum_{y_i\in\mathcal{Y}}[\frac{1}{N_{y_i}}\sum_{n=1}^{N_{y_i}}(1-\cos{(v^{n}_{y_i},s^{n}_{y_i}}))],
   \vspace{\eqsep}
\end{align}
where, $N_{y_i}$ is the \# of training videos of class $y_i$.
Besides, to evaluate the uniformity/separation of semantic clusters, we adopt minimal distances among all clusters to compute dispersion.
Here, we consider all visual features within the same class as a semantic cluster instead of using one single semantic vector $s_{y_i}$.
For example, $\bar{v}_{y_i}$ is the mean of visual features of the class $y_i$, and indicates one semantic cluster.
\begin{align}
    \label{eq:dispersion}
    \text{Dispersion} &= \frac{1}{K}\sum_{y_i\in\mathcal{Y}}\min_{y_k\in\mathcal{Y}\setminus{y_i}}(1-\cos(\bar{v}_{y_i},\bar{v}_{y_k})). 
   \vspace{\eqsep}
\end{align}
The experiments in Sec.~\ref{sec:ablation} show that models tested with higher accuracy preserve the lower closeness and higher dispersion in representations.
We conclude our two metrics can serve as new measurements of model generalizability.

\subsection{Training \& Inference}
{\bf Training}:
We end-to-end train visual and semantic features and jointly learn the contrastive loss $\mathcal{L}_\text{contrast}$ and classification loss $\mathcal{L}_\text{C}$, thus obtaining the following overall loss:
\begin{align}
    \mathcal{L}_\text{AURL} = &\mathcal{L}_\text{S} + \mathcal{L}_\text{US} + \mathcal{L}_\text{C}.
\end{align}
We will justify our end-to-end training is critical for alignment and uniformity properties, and validate our compact model with $\mathcal{L}_\text{AURL}$ outperforms SoTA alternatives.

{\bf Inference:} we train AURL on source dataset $\mathcal{I}$ with $K$ seen classes $\mathcal{Y}=\{y_1, \dots, y_K\}$, and evaluate the model on target dataset $\mathcal{I}^{t}$ with $T$ unseen classes $\mathcal{Y}^{t}=\{y^{t}_1, \dots, y^{t}_T\}$.
In this paper, we follow the strict problem setting in \cite{brattoli2020rethinking}, which requires training classes $\mathcal{Y}$ have no overlap with test classes $\mathcal{Y}^{t}$.
Mathematically, we re-write the requirement as:
\begin{align}
    \forall {y \in \mathcal{Y}}, \; \min_{y^{t} \in \mathcal{Y}^{t}} (1-\cos(c_y, c_{y^{t}})) > \tau,
    \label{eq:label}
    \vspace{\eqsep}
\end{align}
where $c_i$ means Word2vec features of class i, $\tau$ is the distance threshold.
We utilize the Nearest Neighbor Search (NNS) strategy to obtain the final label of query video $I^{t}$:
\begin{align}
 \underset{y^{t} \in \mathcal{Y}^{t}}{\mathrm{argmax}} \cos(f_v(g(I^t)), s_{y^{t}}). 
   \vspace{\eqsep}
   \label{eq:nns}
\end{align}

\subsection{Comparisons with Related Work}
The closest studies to our AURL are SoTA \cite{brattoli2020rethinking}, MUFI \cite{qiu2021boosting}, and ER \cite{chen2021elaborative}.
Table \ref{tab:related} summarizes their similarities and differences.
SoTA only utilizes MSE loss to regularize feature alignment within seen classes, limiting model generalizability.
MUFI and ER both implicitly increase semantic info to improve the generalization.
MUFI trains multi-stream models across multiple datasets.
ER crawls and annotates a number of web words to expand existing class names.
Unlike that MUFI and ER both require extra resources, our AURL is a compact model to utilize the uniformity that helps preserve maximal info of existing features, and introduce a class generator to synthesize more semantics explicitly. 
Even MUFI and ER adopt the supervised contrastive loss, they neglect how alignment and uniformity properties affect ZSVC.
Besides, our AURL follows the strict label requirement (in Eq.~\ref{eq:label}) that classes of training and test sets are far away from each other, which manifests the nature of ZSVC.  
At last, compared with ER that only trains the last fc layers, AURL utilizes a true e2e training strategy that is critical to realize the two properties.
\begin{table}
	\centering\small
	\def\vv{\hspace{2.2pt}}
	\newcommand{\mv}[1]{\vv #1 \vv}
	\caption{Comparisons between AURL and alternative methods.}
	\rowcolors{1}{}{lightgray!80}
	\label{tab:related}
	\begin{tabular}{l*{5}{c}}
	\toprule
		{\bf Methods} & \mv{\bf{ET}} & \mv{\bf{AUL}} & \mv{\bf{ERF}} & \mv{\bf{UCG}} & \mv{\bf{SLR}}\\
	\midrule
		SoTA \cite{brattoli2020rethinking} &$\times$ &$\times$ &$\checkmark$ & $\times$&$\checkmark$\\
    	MUFI \cite{qiu2021boosting} &$\times$ &$\times$ &$\times$ &$\times$&$\times$\\
    	ER \cite{chen2021elaborative}&$\checkmark$ &$\times$ &$\times$ &$\times$&$\times$\\
    	AURL (ours) & $\checkmark$ & $\checkmark$ & $\checkmark$ & $\checkmark$&$\checkmark$\\
	\bottomrule
	\end{tabular}
	\vspace{-6pt}
	\flushleft{*{\bf ET}: end-to-end trainable, {\bf AUL}: alignment-uniformity learning, {\bf ERF}: extra resources free, {\bf UCG}: unseen class generator, {\bf SLR}: strict label requirement.}
	\vspace{\figsep}
\end{table}

\section{Experiments}

\subsection{Settings}
{\bf Datasets:}
We train our AURL on the Kinetics-700 dataset \cite{kay2017kinetics} and evaluate it on UCF101 \cite{soomro2012dataset} and HMDB51 \cite{kuehne2011hmdb} datasets.
The Kinetics-700 provides download links of YouTube videos annotated with 700 categories of human actions.
We collect 555,774 videos using these links.
The UCF101 contains 13,320 videos with 101 actions and the HMDB51 has 6,767 videos annotated with 51 actions.

{\bf Training protocol:}
For fair comparisons with the SoTA \cite{brattoli2020rethinking}, we select the training videos in Kinetics-700 whose classes have non-overlap with UCF101 and HMDB51 as described in Eq. \ref{eq:label}, and set the same $\tau$ = 0.05, thus obtaining 662 classes.
AURL is inductive zero-shot learning, thus does not include any test data during training.

{\bf Evaluation protocol:}
Existing ZSVC methods adopt various evaluation protocols to report experimental results.
For complete comparisons, we perform three protocols: 1, 3, and N test splits.
1 {\em test split} reports an accuracy on all videos of UCF101 or HMDB51 set.
3 {\em test splits} reports an average accuracy by averaging 3 accuracies that are separately evaluated on 3 test sets provided by the UCF101 or HMDB51.
N {\em test splits} also reports the average by averaging N accuracies that are obtained by running N (10 in our method) times testing, in each, m classes are randomly selected (m=50 for UCF101 and m=25 for HMDB51). 

{\bf Implementation details:}
We adopt one or multiple {\em video clips} as one input video of models.
We follow the same SoTA settings \cite{brattoli2020rethinking} (\ie, 1 or 25 video clips and 16 frames/clip with size of $1\times16\times112\times112\times3$) for fair comparisons.
If multiple video clips are used, we take the mean of multiple visual embeddings as the representative embedding.
Besides, if a class name contains multiple words, we average the corresponding Word2vec features to represent the class prototype.
For the AURL architecture, we set feature dimension of the R(2+1)D backbone as 512 (\ie, $g(I)\in\mathbb{R}^{512}$) and dimension of Word2vec as 300 (\ie, $c(\mathcal{Y})\in\mathbb{R}^{K\times300}$), and set the number of nodes in 3-layer MLP of the projector as 2048, 2048, and 2048 separately.
During training, we empirically set $K_u$ as 662, $\lambda$ as 10, $D$ as 662, and $\alpha$ as 0.
We deploy the training on 8 Nvidia Tesla V100 GPUs.
We set batch size as 256 and synchronize all batch normalization across GPUs following \cite{chen2020simple,caron2020unsupervised}.
We implement experiments using PyTorch and Horovod.
SGD is our optimizer and a learning rate of 0.05 with a cosine decay schedule \cite{chen2020simple,loshchilov2016sgdr} is adopted.
Then, we set the weight decay as 0.0001 and the SGD momentum as 0.9.
The number of training iterations is 58,500 which takes 45 hours.

\begin{figure*}
    \centering
    \includegraphics[width=0.95\textwidth]{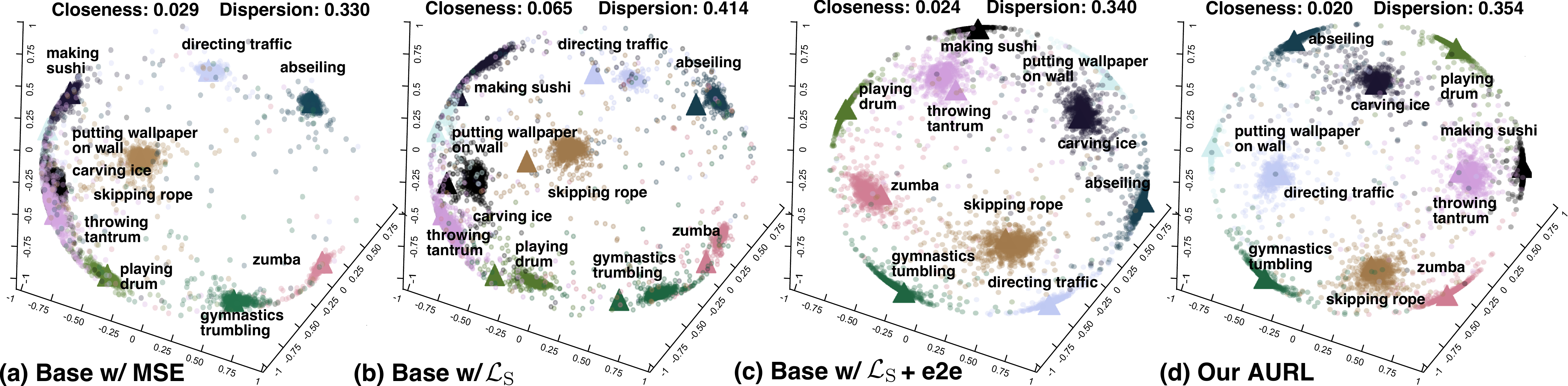}
    \caption{
    {\bf Ablations:} The representations of ablations w/ MSE loss, our $\mathcal{L}_\text{S}$, e2e training, and the AURL.
    \protect\scalebox{1}{\textbullet} and \protect$\bigtriangleup$ represent visual and semantic features separately; colors are for different classes.
    Here, we randomly sample 10 classes from Kinetics-700 for visualization.
    }
    \label{fig:loss}
    \vspace{\figsep}
\end{figure*}

\subsection{Ablation Study}
\label{sec:ablation}
To analyze AURL, we performed extensive ablations that were trained on the Kinetics-700 and evaluated on UCF101 and HMDB51 using 1 {\em video clip} and 1 {\em test split} of evaluation protocols.
Table~\ref{tab:ablations} summarizes the quantitative results.
Fig.~\ref{fig:loss} visualizes the visual-semantic representation of ablations by sampling 10 classes from Kinetics-700 dataset and setting the features as 3-D for better visualization.
We will justify:
(1) our model that preserves alignment-uniformity properties performs better than the SoTA method \cite{brattoli2020rethinking} that focuses on alignment only;
(2) end-to-end (e2e) training is critical to realize the two properties;
(3) our AURL involving the class generator performs the best.
From the justifications, we will show that our closeness and dispersion metrics can serve as new measurements of model generalizability.
Here, we take the architecture of the SoTA \cite{brattoli2020rethinking} as our {\bf Base} model (\ie, base backbone + fc only for video parts).

Per (1), we compare {\bf Base w/ MSE} (\ie, the SoTA) and {\bf Base w/ $\mathbf{\mathcal{L}_\text{S}}$}.
From the accuracy in Table~\ref{tab:ablations}, {\bf Base w/ $\mathbf{\mathcal{L}_\text{S}}$} largely improves the results by (14.8\%, 35.1$\rightarrow$40.3) on UCF101 and by (13.6\%, 21.3$\rightarrow$24.2) on HMDB.
Comparing the learned representation in Fig.~\ref{fig:loss} (a) {\bf w/ MSE}  and (b) {\bf w/ $\mathbf{\mathcal{L}_\text{S}}$} , we observe the semantic clusters of (b) spread more than (a), but the alignment within classes gets worse, for example, visual and semantic features are not calibrated for classes ``skipping rope" and ``abseiling".
Similarly, we find the same trend in closeness and dispersion metrics shown in Fig.~\ref{fig:loss} and Table~\ref{tab:ablations}, where closeness gets worse (0.029$\rightarrow$0.065, 0.30$\rightarrow$0.45) but dispersion becomes much better (0.330$\rightarrow$0.414, 0.09$\rightarrow$0.29).
We can see our {\bf Base w/ $\mathbf{\mathcal{L}_\text{S}}$} presesrving higher uniformity in the trained representation can achieve better generalization when making inference on the test set, even scarify a little alignment. 

Per (2), we involve {\bf e2e} training strategy (\ie, base backbone + video projector for video parts; word projector for semantic parts) to the {\bf Base w/ $\mathbf{\mathcal{L}_\text{S}}$}, and get the {\bf Base w/ $\mathbf{\mathcal{L}_\text{S}}$ + e2e}, which further improves the accuracy from 40.3 to 43.2 on UCF-101 and from 24.2 to 26.2 on HMDB.
Not surprisingly, we observe the alignment is tuned better and uniformity is maintained in good quality, thus obtaining a better trade-off.
Referring to Fig.~\ref{fig:loss} (b) and (c), we see {\bf Base w/ $\mathbf{\mathcal{L}_\text{S}}$ + e2e} encourages better uniformity that semantic clusters are relatively distributed uniformly across the hypersphere while achieves a satisfying alignment that visual and semantic features are apparently aligned (see classes ``skipping rope" and ``abseiling" again for comparisons).
The similar trends also occur in closeness and dispersion metrics, \ie, (0.024 and 0.340; 0.30 and 0.29) in Fig.~\ref{fig:loss} and Table~\ref{tab:ablations}.
We conclude that e2e is critical for adjusting features to meet the regularizations of alignment and uniformity.

Per (3), we apply {\bf $\mathbf{\mathcal{L}_\text{US}}$} to unseen classes coupling with the class generator ({\bf CG}), \ie, our {\bf AURL}.
Compared {\bf AURL} with {\bf Base w/ $\mathbf{\mathcal{L}_\text{S}}$ + e2e}, {\bf AURL} steadily improves 2.8\% on UCF and 4.6\% on HMDB, achieving the best accuracy.
Quantitatively, closeness and dispersion reach the best scores, such as (0.29, 0.32)  in Table~\ref{tab:ablations} and (0.020, 0.354) in Fig.~\ref{fig:loss}.
From the representation of Fig.~\ref{fig:loss} (d), we see the semantic clusters cover most regions of the hypersphere, which improves the possibility that unseen features fall around existing points, thus bringing a better generalization.
Furthermore, we remove {\bf CG} from AURL (\ie, {\bf AURL w/o CG}) to validate the effectiveness of the class generator.
Comparing {\bf AURL} and {\bf AURL w/o CG}, we find that the performances of {\bf AURL w/o CG} on UCF and HMDB both decrease, and the accuracy on HMDB even degrades lower than {\bf Base w/ $\mathbf{\mathcal{L}_\text{S}}$ + e2e}.
Thus, we conclude the {\bf CG} is a critical module to enhance the generalization.

Last but not least, from the above justifications, we summarize that our models consistently improve the accuracy by involving the proposed modules;
closeness/dispersion measured on the learned representations have agreements with the accuracy evaluated on test sets, providing model evaluations even prior to making inference.

\def\vv{\hspace{3pt}}
\begin{table}[t]
\small
	\centering
	\caption{
	         Ablations of our modules using 1 video clip under the {\em 1 test split} protocol ($\tau$=0.05).
	         {\red{Red}} numbers indicate the best. Closeness$\downarrow$ better, dispersion$\uparrow$ better, and top-1 accuracy $\uparrow$ better.}
	\label{tab:ablations}
	\begin{tabular}{@{}l@{\vv}|c@{\vv}c@{\vv}c@{\vv}|c@{\hspace{7pt}}c@{\hspace{7pt}}|c@{\hspace{5pt}}c@{}}
	\toprule
		{\bf Method} & {\bf $\mathbf{\mathcal{L}_\text{S}}$} &\bf{e2e}   & \shortstack{{\bf $\mathbf{\mathcal{L}_\text{US}}$}\\ +{\bf CG}} &\shortstack{\bf{Clo-} \\ \bf{se.}}&\shortstack{\bf{Dis-} \\ \bf{per.}} &\shortstack{\bf{UCF} \\ \bf{top-1}}& \shortstack{\bf{HMDB} \\ \bf{top-1}}\\
	\midrule
		\bf{Base w/ MSE} &&  & &0.30&0.09&35.1&21.3\\
		\bf{Base w/ $\mathbf{\mathcal{L}_\text{S}}$} &\checkmark&&&0.45&0.29&40.3&24.2\\
		\bf{Base w/ $\mathbf{\mathcal{L}_\text{S}}$ + e2e} &\checkmark& \checkmark &&0.30&0.29&43.2&26.2\\
		\rowcolor{gray!10}
		\bf{AURL (ours)} &\checkmark&\checkmark &\checkmark &{\red 0.29}&\red{0.32}&\red{44.4}&\red{27.4}\\
		\bf{AURL w/o CG} &\checkmark&\checkmark &$\bcancel{\text{CG}}$ & 0.33 &0.32&43.7&25.8\\
	\bottomrule
	\end{tabular}
	\vspace{\figsep}
\end{table}

\subsection{Comparisons with the Closest SoTA}
The closest SoTA to our AURL is the recent work \cite{brattoli2020rethinking}, which utilizes a compact model that achieves the SoTA results even under a strict setting (\ie, Eq.~\ref{eq:label}).
Table \ref{tab:related} summarizes the similarity and difference between the SoTA and our AURL.
Table \ref{tab:closest} shows the comprehensive comparisons quantitatively.
We reported the SoTA results using the same settings and the authors’ released code.
For comprehensive comparisons, we include various evaluation protocols including Pre-training, Video clips, and Test splits.
Pre-training means that SoTA fine-tunes the pre-trained models on the SUN dataset \cite{xiao2010sun}.
From the comparisons, we see our AURL consistently surpasses the SoTA under each evaluation protocol.
Specifically, the smallest improvements happen at (25 Video clips, 1 Test splits) by (17.6, 16.5)\% improvements on UCF top-1 and HMDB top-1, and the largest comes at (1 Video clip, 10 Test split) by (28.1, 27.0)\% increases on UCF top-1 and HMDB top-1.
{\bf To conclude, AURL outperforms the SoTA by a large margin.}

\def\vv{\hspace{3.1pt}}
\begin{table}[t]
\small
	\centering
	\caption{
	        Comparisons with the closest SoTA \cite{brattoli2020rethinking} on both UCF and HMDB datasets.
	        {\color{red}{Red}} numbers indicate the best.}
	\label{tab:closest}
	\begin{tabular}{@{}l@{\hspace{2pt}}|c@{\vv}c@{\vv}c@{\vv}|c@{\hspace{5.2pt}}c@{\hspace{5.2pt}}c@{\vv}c@{}}
	\toprule
		\rb{4}{\ {\bf Method}\ } & {\shortstack{{\bf Pre-}\\{\bf training}}} &{\shortstack{{\bf Video}\\{\bf clips}}} &{\shortstack{{\bf Test}\\{\bf splits}}} &{\shortstack{{\bf UCF}\\{\bf top-1}}}&{\shortstack{{\bf UCF}\\{\bf top-5}}}&{\shortstack{{\bf HMDB}\\{\bf top-1}}}&{\shortstack{{\bf HMDB}\\{\bf top-5}}}\\
	\midrule
	    &~ &1 &10 &43.0 &68.2 &27.0&54.4\\
	    \multirow{-2}{*}{SoTA}&\checkmark &1 &10 &45.6 &73.1 &28.1 &51.8  \\
	    \rowcolor{gray!10}
        AURL & &1 &10 & {\red 55.1} &{\red 79.3} &{\red 34.3} &{\red 65.1}\\
	\midrule	
	    &  &25 &10 &48.0 &74.2 &31.2 &58.3\\
	    \multirow{-2}{*}{SoTA} &$\checkmark$ &25 &10 &49.2 &77.0 &32.6 &57.1 \\
	    \rowcolor{gray!10}
        AURL & &25 &10 &{\red 58.0} &{\red 82.0} &{\red 39.0} &{\red 69.5}\\
	\midrule	
		& &1 &1 &35.1 &56.4 &21.3 &42.2\\
	    \multirow{-2}{*}{SoTA}&\checkmark &1 &1 &36.8 &61.7 &23.0 &41.3\\
	    \rowcolor{gray!10}
        AURL & &1 &1 &{\red 44.4} &{\red 70.0} &{\red 27.4} &{\red 53.2}\\
	\midrule	
	    &  &25 &1 &37.6 &62.5 &26.9 &49.8\\
	    \multirow{-2}{*}{SoTA} &$\checkmark$ &25 &1 &39.8 &65.6 &27.2 &47.4\\
	    \rowcolor{gray!10}
        AURL & &25 &1 &{\red 46.8} &{\red 73.1} &{\red 31.7} &{\red 58.9}\\
	\bottomrule
	\end{tabular}
	\vspace{\figsep}
\end{table}

\subsection{Comparisons with the Alternatives}
Table \ref{tab:comparisons} shows the comparisons with the alternatives.
The SoTA \cite{brattoli2020rethinking} and our AURL with $\star$ mean the two methods follow the strict label requirement in Eq.\ref{eq:label}.
From the results, we observe that our AURL surpasses all the alternatives in various challenging situations.
Specifically, we summarize the challenges:
first, fewer test splits are harder testing situations, \eg, for SAOE \cite{mettes2017spatial}, 3 \vs 10 splits corresponds to 32.8 \vs  40.4 on UCF;
second, strict label requirement ($\star$) serves more difficult situation, \eg, our AURL$^\star$ w/ 10 (the more) test splits achieves even worse results than AURL w/ 3 splits;
third, some methods acquire extra training datasets (\eg, Kinetics + extra 5 datasets trained in MUFI \cite{qiu2021boosting}), additional semantic classes (\eg, web words used in ER \cite{chen2021elaborative} ), and even training videos sampled from the same domain as the test set (\eg, tr/te are both UCF or HMDB in TARN \cite{bishay2019tarn}, Act2Vec \cite{hahn2019action2vec}, PSGNN \cite{gao2020learning}, and ER \cite{chen2021elaborative}), which provide more difficulties to be competed against for other methods.
Correspondingly, we find the superiority of our AURL as below:
(1) AURL w/ 1 (the fewest) test split outperforms most methods w/ 3 or 50 splits, \eg, (46.8, 31.7) of AURL \vs (36.3, -) of OPCL and (43.0, 32.6) of PSGNN on (UCF, HMDB) dataset;
(2) AURL$^\star$ w/ more strict requirements but w/o extra datasets competes against all the SoTA alternatives, \eg, (58.0, 39.0) of AURL \vs (51.8, 35.3) of ER, and (56.3, 31.0) of MUFI on (UCF, HMDB) dataset.
{\bf To sum up, our AURL reaches the new SoTA in ZSVC}.

Finally, we conduct AURL$^\star$ with pre-extracted features (trained only on video and word projectors).
We observe AURL w/ pre-extracted features achieves comparable performance with the e2e AURL -- (59.5, 38.2) \vs (58.0, 39.0) on UCF and HMDB.
This suggests AURL can achieve high performance without carefully finetuning video features.

{\bf Limitations and possible solutions.}
Even our AURL achieves promising results, there are still two problems to be concerned.
(1) Uniformity of visual features within classes could be included to further increase info-preserving, introducing contrastive learning between videos may be a possible solution.
(2) It will be helpful to study how the class generator affects the overall optimization during training.

\def\vv{\hspace{5pt}}
\begin{table}[t]
\small
	\centering
	\caption{
	        Comparisons with SoTA alternatives on both UCF and HMDB datasets.
	        Results of alternatives were obtained from original papers, and the higher, the better.
	        {\red{Red}} and {\blue{blue}} numbers indicate the best and second best.
	        $\star$ means using $\tau$=0.05 in Eq.\ref{eq:label}.}
	\label{tab:comparisons}
	\begin{tabular}{@{\hspace{10pt}}l@{\vv}|c@{\vv}|c@{\vv}c@{\vv}|c@{\vv}c@{\vv}}
	\toprule
		{\bf Method} &{\shortstack{{\bf Test}\\{\bf splits}}}&{\shortstack{{\bf Train}\\{\bf dataset}}}&{\shortstack{{\bf UCF}\\{\bf top-1}}} &{\shortstack{{\bf Train}\\{\bf dataset}}}&{\shortstack{{\bf HMDB}\\{\bf top-1}}}\\
	\midrule
	    SoTA$^{\star}$ \cite{brattoli2020rethinking} &1 &Kinetics&37.6 &Kinetics&26.9\\
	\rowcolor{gray!10}
	    AURL$^{\star}$  & 1 &Kinetics&{\red 46.8} &Kinetics&{\red 31.7}\\
	\midrule	
	    Obj2act \cite{jain2015objects2action}&3 &-&30.3 &-&15.6\\
		SAOE \cite{mettes2017spatial}&3 &-&32.8 &-&-\\
		OPCL \cite{gao2020learning}&3 &-&36.3 &-&-\\
		MUFI\cite{qiu2021boosting} &3 &Kinetics+&{\blue 56.3} &Kinetics+&{\blue 31.0}\\
	\rowcolor{gray!10}
		AURL  &3 &Kinetics&{\red 60.9} &Kinetics&{\red 40.4}\\   
	\midrule
	    TARN \cite{bishay2019tarn} &30 &UCF&23.2 &HMDB&19.5\\
		Act2Vec \cite{hahn2019action2vec}&- &UCF&22.1 &HMDB&23.5\\
		SAOE\cite{mettes2017spatial} &10 &-&40.4 &-&-\\
		PSGNN \cite{gao2020learning} &50 &UCF&43.0 &HMDB&32.6\\
		OPCL \cite{gao2020learning}&10 &-&47.3 &-&-\\
		SoTA$^{\star}$\cite{brattoli2020rethinking} &10 &Kinetics&48.0 &Kinetics&{32.7}\\
		DASZL \cite{kim2021daszl} &10 &-&{48.9} &-&-\\
		ER \cite{chen2021elaborative} &50 &UCF&\blue{51.8}&HMDB&\blue{35.3}\\
	\rowcolor{gray!10}
		AURL$^{\star}$ &10 &Kinetics&{\red 58.0} &Kinetics&{\red 39.0}\\
	\bottomrule
	\end{tabular}
	\vspace{\figsep}
\end{table}

\section{Conclusion}
This paper learns representation awareness of both alignment and uniformity properties for seen and unseen classes.
We reformulate a supervised contrastive loss to jointly align visual-semantic features and encourage semantic clusters to distribute uniformly.
To explicitly synthesize features of unseen semantics, we propose a class generator that performs feature transformation on features of seen classes.
Besides, we introduce closeness and dispersion to quantify the two properties, providing new measurements for generalization.
Extensive ablations justify the effectiveness of each module in our model.
Comparisons with the SoTA alternatives validate our model reaches the new SoTA results.

{\bf Acknowledgement.}
Research was supported by Natural Science Foundation of China under grant 62076036.

{\small
\bibliographystyle{ieee_fullname}
\bibliography{aucrl}
}

\end{document}